\documentclass[runningheads]{llncs}
\usepackage[T1]{fontenc}
\usepackage{hyperref}
\usepackage{graphicx}
\usepackage{caption}
\usepackage{graphicx,verbatim}
\usepackage{amsmath}
\usepackage{amssymb}
\usepackage{xcolor}
\begin{document}
\title{MvHo-IB: Multi-View Higher-Order Information Bottleneck for Brain Disorder Diagnosis}
%
\author{Kunyu Zhang\inst{1} \and
Qiang Li\inst{2} \and
Shujian Yu\inst{3,4}\orcidID{0000-0002-6385-1705}}
\authorrunning{K. Zhang et al.}
\institute{International College, Zhengzhou University, 450000 Zhengzhou, China\\
\email{kunyu.zky@gmail.com} \and
Tri-Institutional Center for Translational Research in Neuroimaging and Data Science (TReNDS), Georgia State, Georgia Tech, and Emory University, Atlanta, GA 30303, USA
\and
Department of Computer Science, Vrije Universiteit Amsterdam, 1081 HV Amsterdam, The Netherlands\\
\and
Department of Physics and Technology, UiT - The Arctic University of Norway, 9019 Troms\o, Norway \\
\email{yusj9011@gmail.com} }

\maketitle           
\begin{abstract}
Recent evidence suggests that modeling higher-order interactions (HOIs) in functional magnetic resonance imaging (fMRI) data can enhance the diagnostic accuracy of machine learning systems. However, effectively extracting and leveraging HOIs remains a significant challenge. In this paper, we propose MvHo-IB, a novel multi-view learning framework that seamlessly integrates pairwise interactions and HOIs for diagnostic decision-making while automatically compressing task-irrelevant redundant information. Our approach introduces several key innovations: (1) a principled framework combining $\mathcal{O}$-information from information theory with the recently developed matrix-based R\'enyi’s $\alpha$-order entropy functional estimator to quantify and extract HOIs, (2) a purpose-built Brain3DCNN encoder designed to effectively utilize these interactions, and (3) a novel multiview learning information bottleneck objective to enhance representation learning. Experiments on three benchmark fMRI datasets demonstrate that MvHo-IB achieves state-of-the-art performance, outperforming existing methods, including modern hypergraph-based techniques, by significant margins. The code of our MvHo-IB is available at \url{https://github.com/zky04/MvHo-IB}.
\end{abstract}
\section{Introduction}
Mental disorders exhibit complex neural signatures, making precise neurobiological characterization challenging. Resting-state functional magnetic resonance imaging (rs-fMRI) has emerged as a cornerstone for machine learning-based diagnostic frameworks in mental disorders~\cite{marshall2020hidden}. With the advent of deep learning, researchers have developed more sophisticated models to analyze brain networks. Convolutional neural networks primarily captured localized functional connectivity (FC) patterns~\cite{kawahara2017brainnetcnn}, while graph neural networks (GNNs) further advanced whole-brain analysis by leveraging the complex relational structure of brain networks~\cite{martensson2018stability}.

However, existing deep learning approaches face two key limitations. First, they predominantly rely on correlations or partial correlations to characterize linear and pairwise FC between brain regions. This fundamentally oversimplifies the role of higher-order interactions (HOIs), which are essential for understanding complex cognitive processes~\cite{prado2022neurobiology}. Second, they overlook the impact of noisy or spurious interactions (i.e., connections influenced by measurement noise or patient-specific artifacts) on the final decision, which may degrade generalization performance and lead to unreliable predictions.

Growing evidence suggests that functional HOIs involving more than two brain regions play a crucial role in neural computation. To model these complex interactions, recent computational approaches have leveraged hypergraph theory~\cite{feng2019hypergraphneuralnetworks,Wang2024}. Specifically, these methods employ a dual representation scheme, where hyper-nodes map to anatomically segregated brain regions and hyper-edges explicitly encode multivariate functional dependencies~\cite{Wang2024}. This formulation enables the modeling of concurrent activation patterns across three or more regions, offering a neurobiologically plausible representation of network-level dynamics. However, while hypergraph-based methods offer theoretical advantages for HOI characterization, their practical deployment faces significant bottlenecks. Specifically, hypergraph approaches require manually constructing high-order networks by selecting similarity metrics and pruning rules. The resulting hyperedges, each connecting an arbitrary number of brain regions, merely indicate that these regions are connected, without revealing how they share information (e.g., redundantly or synergistically).

Rather than relying on hypergraphs to model HOIs in a \emph{model-driven} manner, we propose a \emph{data-driven} approach by leveraging $\mathcal{O}$-information~\cite{Rosas2019QuantifyingHI,varley2023multivariate} from information theory to capture HOIs.
Unlike hyperedges, $\mathcal{O}$-information provides a single signed measure that indicates if a set of brain regions generates genuinely new joint information (negative value, synergy-dominated) or primarily reflects repeated signals (positive value, redundancy-dominated). That is, $\mathcal{O}$-information not only captures whether regions are connected, as in the case of hyperedges, but also provides a fine-grained quantification of the nature of their interaction. Moreover, constructing HOIs with $\mathcal{O}$-information does not require manually selecting similarity metrics or applying pruning rules that may introduce bias.

Our contributions can be summarized as follows:
\begin{itemize}
    \item We develop MvHo-IB, a novel multi-view learning framework that seamlessly integrates both \emph{nonlinear} pairwise interactions and HOIs while simultaneously eliminating redundant information to enhance predictive performance. 

    \item We propose leveraging $\mathcal{O}$-information to capture HOIs and introduce the matrix-based Rényi’s $\alpha$-order entropy estimator~\cite{8787866} for its computation. Additionally, we develop Brain3DCNN, a specialized architecture that exploits the topological locality of structural brain networks to enhance $\mathcal{O}$-information representation learning.

    \item Our MvHo-IB outperforms eight widely used brain network classification methods, demonstrating strong generalization across three datasets while providing clinically interpretable insights that align with clinical evidence.
\end{itemize}

\section{Information Bottleneck in Brain Disorder Diagnosis}
The IB principle is a framework for extracting the most relevant information from an input variable $X$ for predicting an output variable $Y$. It operates by identifying a ``bottleneck'' variable $Z$ that maximizes its predictive power to $Y$, as expressed by the mutual information $I(Y;Z)$, while imposing some constraints on the amount of information it carries about $X$, formulated as $I(X;Z)$:
\begin{equation}
\mathcal{L}_{\text{IB}} = I(Y; Z) - \beta I(X; Z),
\end{equation}
where \(\beta > 0\) is a Lagrange multiplier.

Recent studies have employed the IB principle to enhance both interpretability and generalization in graph-structured data. For example, the dynamic graph attention information bottleneck framework~\cite{DONG2024} refines raw brain graphs by optimizing graph connectivity and reducing noise, enhancing effective feature aggregation.

The Subgraph Information Bottleneck (SIB)~\cite{yu2021recognizing} focuses on automatically extracting a predictive subgraph to explain the final decision, thereby enhancing interpretability. Furthermore, BrainIB~\cite{BrainIB} stabilizes the training of SIB by utilizing the matrix-based R\'enyi's $\alpha$-order entropy functional, applying this refined framework to the diagnosis of mental disorders.

\section{Methodology}

Consider a dataset of brain signal recordings $\{X^i, Y^i\}_{i=1}^N$, where each recording $X^i \in \mathbb{R}^{C \times T}$ represents the raw blood-oxygen-level-dependent (BOLD) signal, detected in fMRI, for the $i$-th patient. Here, $C$ denotes the number of channels (e.g., $116$ for AAL atlas~\cite{Tzourio02} or $105$ for ICA-driven brain network template~\cite{Iraji2022CanonicalAR}) and $T$ indicates the signal duration. We further use subscripts to denote the channel index of BOLD signal. Specifically, \(X^j_i\) represents the $1$D signal from the \(i\)-th ($1\leq i\leq C$) brain region for the \(j\)-th ($1\leq j\leq N$) patient. Our objective is to develop a classifier that maps raw brain signal $X$ to its label $Y$.

\subsection{Multi-view Information Bottleneck}

\begin{figure}[h]
    \centering
    \includegraphics[width=0.9\textwidth, height=0.3\textheight]{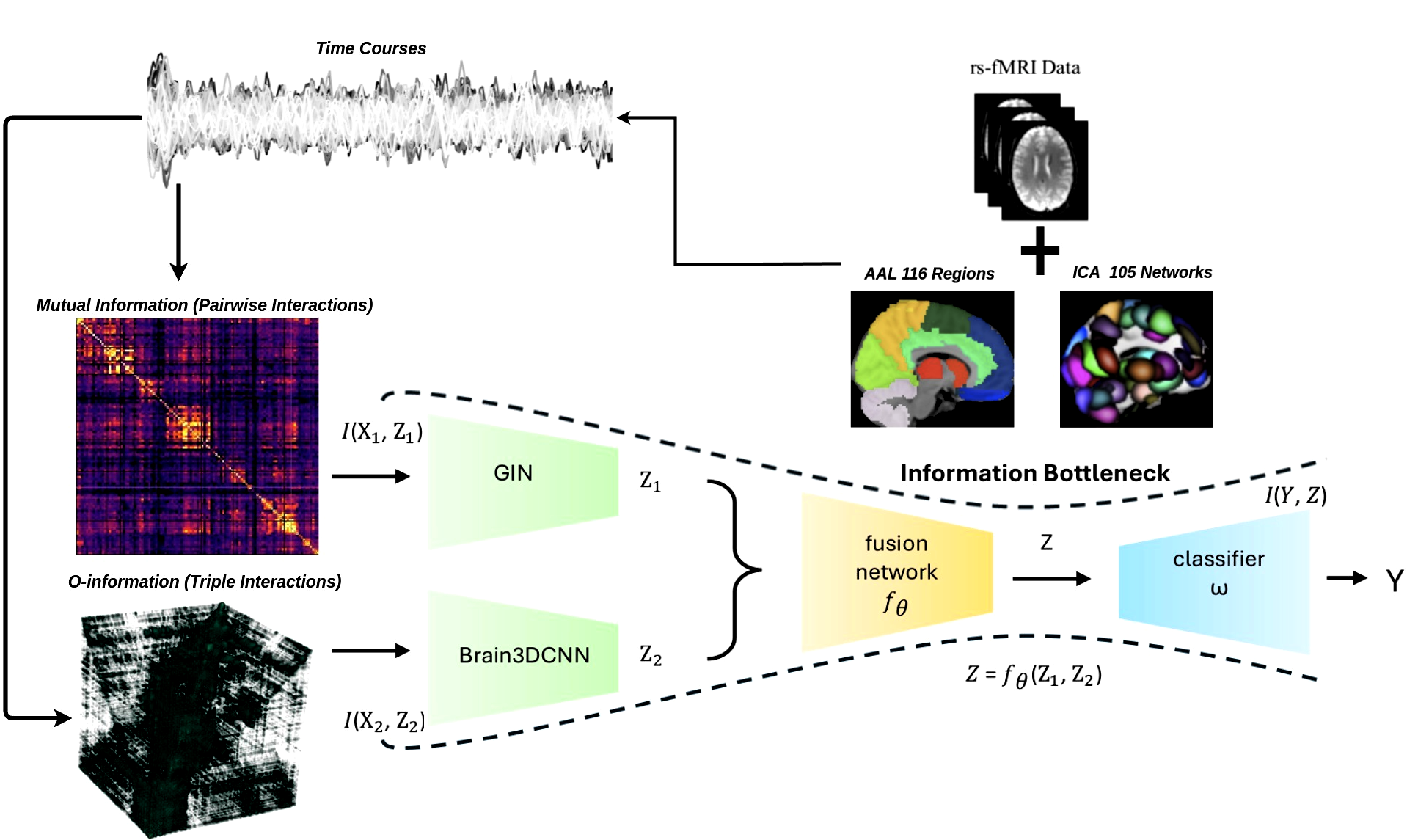}
    \caption{Illustration of MvHo-IB. The time courses are extracted from fMRI using the automated anatomical labeling (AAL) template~\cite{tzourio2002automated}. Then, functional connectivity patterns are estimated before being fed into a multi-view framework. The framework learns a joint representation \(Z = f_{\theta}(Z_1, Z_2)\) by balancing the maximization of \(I(Y; Z)\) with the minimization of \(I(X_1; Z_1) + I(X_2; Z_2)\), where the first view input is the mutual information matrix (pairwise interactions) and the second view input is the \(\mathcal{O}\)-information 3D tensor (triple interactions).}
    \label{fig:mvframework}
\end{figure}

Figure~\ref{fig:mvframework} illustrates the architecture of MvHo-IB. For each participant, we derive two types of brain representations: a $C\times C$ matrix that encodes all pairwise FCs and a $C\times C\times C$ 3D tensor that captures all three-way interactions, representing the interdependencies among any triplet of brain regions.
We treat the raw 2D and 3D representations as complementary views of the raw BOLD signals and propose using a multi-view learning framework to learn a compact and informative joint representation $z$. To achieve this, MvHo-IB consists of two encoders \(h_{\phi_1}\) (a Graph Isomorphism Network or GIN to learn $z_1$) and \(h_{\phi_2}\) (a purpose-designed Brain3DCNN that will be detailed in Section 3.3 to learn $z_2$), a feature fusion network \(f_\theta\) to integrate $z_1$ and $z_2$ into $z$, and a nonlinear classifier \(h_w\). 

Thus, the overall objective of MvHo-IB is formulated as:
\begin{equation}
\begin{aligned}
&\mathop{\arg\max}_{\phi_1, \phi_2, \theta, w} \Bigl( I(Y; Z) - \Bigl( \beta_1 I(X_1; Z_1) + \beta_2 I(X_2; Z_2) \Bigr) \Bigr), \\
&\hspace{2.3cm} \text{s.t.} \quad Z = f_\theta(Z_1, Z_2),
\end{aligned}
\label{eq3}
\end{equation}
where \(\beta_1\) and \(\beta_2\) are the regularization coefficients for views 1 and 2, respectively.

The prediction term \(I(Z; Y)\) is lower-bounded as follows~\cite{Kolchinsky_2019,alemi2019deepvariationalinformationbottleneck}:
\begin{equation}
\label{eq4}
    I(Z; Y) \geq H(Y) + \mathbb{E}_{P(Z, Y)}\bigl[\log P(Y \mid Z)\bigr],
\end{equation}
which essentially optimizes the cross-entropy loss \(\operatorname{CE}(Y, \hat{Y}) = -\mathbb{E}_{P(Z, Y)}\bigl[\log P(Y \mid Z)\bigr]\), since $H(Y)$ is a constant only depends on input data. On the other hand, for deterministic encoders, $I(X;Z)=H(Z)$ as the mapping uncertainty $H(Z|X)=0$~\cite{michael2018on,strouse2017deterministic}. Consequently, Eq.~\ref{eq3} is reformulated as:
\begin{equation}
\begin{aligned}
&\mathop{\arg\min}_{\phi_1, \phi_2, \theta, w} \Bigl( \operatorname{CE}(Y, \hat{Y}) + \beta_1 H(Z_1) + \beta_2 H(Z_2) \Bigr), \\
&\hspace{2.3cm} \text{s.t.} \quad Z = f_\theta(Z_1, Z_2).
\end{aligned}
\label{eq5}
\end{equation}

\subsection{$\mathcal{O}$-Information and Matrix-based Entropy Functional}
We propose utilizing $\mathcal{O}$-information~\cite{varley2023multivariate,Rosas2019QuantifyingHI} to capture HOI among any triplet of brain regions. Formally, the $\mathcal{O}$-information is defined as the difference between the total correlation (TC)~\cite{TC} and the dual total correlation (DTC)~\cite{DTC}, both of which are nonlinear multivariate dependence measures~\cite{yu2021measuring}. The $\mathcal{O}$-information for BOLD signals from the $i$-th, $j$-th, and $k$-th brain regions is defined as:
\begin{equation}
\mathcal{O}(X_i, X_j, X_k) = T(X_i, X_j, X_k) - D(X_i, X_j, X_k),
\end{equation}
if \(\mathcal{O} > 0\), the triplet is redundancy-dominated, if \(\mathcal{O} < 0\), synergy dominates~\cite{varley2023multivariate}.

The first term TC \( T(X_i, X_j, X_k) \) is defined as the Kullback-Leibler (KL) Divergence between the joint distribution $p(X_i,X_j,X_k)$ and product of marginals $p(X_i)p(X_j)p(X_k)$, which can be further decomposed as:
\begin{equation}
T(X_i, X_j, X_k) =  H(X_i) + H(X_j) + H(X_k) - H(X_i, X_j, X_k),
\end{equation}
where \( H\) denotes entropy or joint entropy, which can be elegantly estimated with the matrix-based R\'enyi’s \(\alpha\)-order entropy functional~\cite{yu2021measuring}.

Analogously with TC, the DTC \( D(X_i, X_j, X_k) \) is defined as~\cite{DTC}:
\begin{equation}
D(X_i, X_j, X_k) = H(X_i, X_j, X_k) - H(X_i \,|\, X_j,X_k) - H(X_j \,|\, X_i,X_k) - H(X_k \,|\, X_i,X_j),
\end{equation}
where \( H(\cdot | \cdot) \) is the conditional entropy.

\subsection{Brain3DCNN}

\subsubsection{3D Edge-to-Edge (E2E) Layer}

\begin{figure}[htbp]
    \centering
    \includegraphics[width=\textwidth,height=3cm]{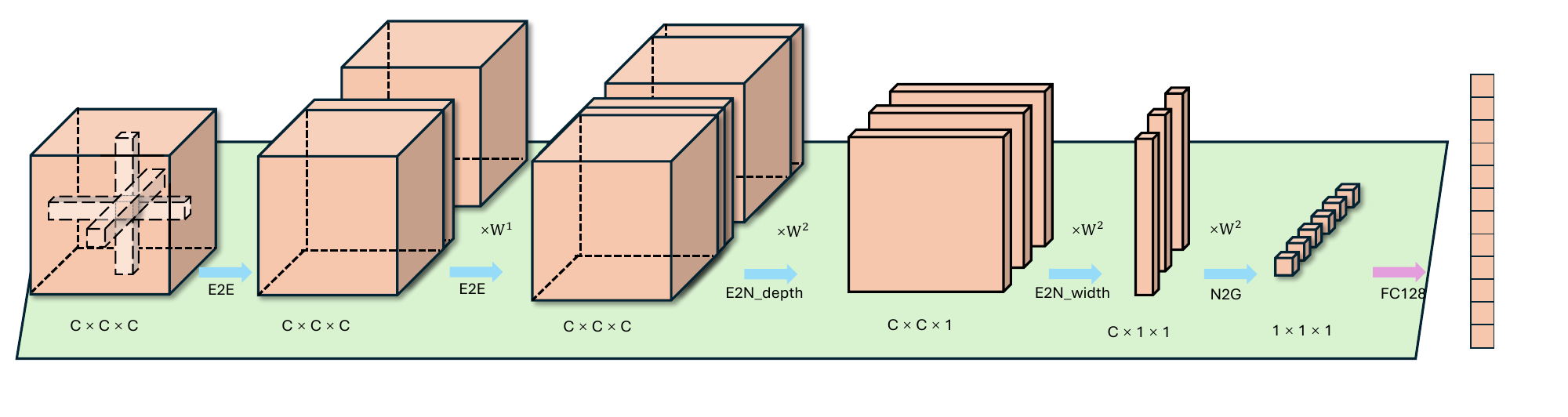}
    \caption{Each block represents the input/output of filter layers. The brain network adjacency matrix (leftmost block) undergoes E2E convolution, followed by E2N filtering to aggregate edge weights per region. The N2G layer integrates node responses, and fully connected layers refine features for the final prediction.}
    \label{fig:pdfimage}
\end{figure}

Inspired by BrainNetCNN~\cite{kawahara2017brainnetcnn}, our 3D E2E layer processes the 3D tensor
by aggregating information from connected edges in a trident kernel along three spatial dimensions (see Figure~\ref{fig:pdfimage}). 
Formally, let $\mathbf{O}^{(\ell)} \in \mathbb{R}^{M^\ell \times C \times C \times C}$ denote all 
feature maps at the $\ell$-th layer, extracted from $M^\ell$ convolutional kernels. For the first input layer, $\mathbf{O}^{1} \in \mathbb{R}^{C\times C\times C}$ is the estimated $\mathcal{O}$-information tensor.
The output $\mathbf{O}^{(\ell+1,n)}$ for the $n$-th feature map ($1\leq n\leq M^{\ell+1}$) at the layer $(\ell+1)$ is given by:
\begin{equation}
O^{(\ell+1,n)}_{i,j,k} \;=\;
\sum_{m=1}^{M^\ell} 
\sum_{c=1}^{C}
\Bigl[
r^{(\ell,m,n)}_{d}\,O^{(\ell,m)}_{i-c,\,j,\,k}
\;+\;
d^{(\ell,m,n)}_{d}\,O^{(\ell,m)}_{i,\,j-c,\,k}
\;+\;
e^{(\ell,m,n)}_{d}\,O^{(\ell,m)}_{i,\,j,\,k-c}
\Bigr],
\label{eq:e2e_3d}
\end{equation}
where $r^{(\ell,m,n)}_{d}$, $d^{(\ell,m,n)}_{d}$, and $e^{(\ell,m,n)}_{d}$ are learnable weights. 
Here, $C$ denotes the spatial extent in each dimension 
(e.g., $116$ for AAL atlas~\cite{Tzourio02} or $105$ for ICA-driven brain network template~\cite{Iraji2022CanonicalAR}).

\subsubsection{3D Edge-to-Node (E2N) Layer}
The 3D E2N layer aggregates the edge-focused representation into a 2D node feature map. 
Let $\mathbf{O}^{(\ell,m)} \in \mathbb{R}^{I \times J \times K}$ again be the input 
feature map; the E2N layer output $a_i^{(\ell+1,n)}$ for node $i$ under the $n$-th feature map at layer 
$(\ell+1)$ is:
\begin{equation}
a_i^{(\ell+1,n)} \;=\; 
\sum_{m=1}^{M^\ell} \sum_{k=1}^{K}
\Bigl[
r_k^{(\ell,m,n)}\,O^{(\ell,m)}_{i,k} 
\;+\;
d_k^{(\ell,m,n)}\,O^{(\ell,m)}_{k,i}
\Bigr],
\label{eq:e2n_3d}
\end{equation}
where $r_k^{(\ell,m,n)}$ and $d_k^{(\ell,m,n)}$ are the learnable weights. We apply this 3D-to-2D compression
sequentially along spatial dimensions ($I$, $J$, or $K$) to reduce the tensor size while preserving
key topological relationships between brain regions.

\subsubsection{3D Node-to-Graph (N2G) Layer}
The 3D N2G layer aggregates node features into a scalar. It performs convolution across nodes, capturing spatial relationships in depth, height, and width.

\section{Experiments and Results}
\subsection{Datasets and Experimental Settings}
We evaluate our method on three real-world fMRI datasets. The first dataset, from the UCLA Consortium for Neuropsychiatric Phenomics~\cite{Gorgolewski17}, includes schizophrenia (SZ, $n=50$) and normal controls (NC, $n=114$). The second, from the Alzheimer’s Disease Neuroimaging Initiative (ADNI)~\cite{Kuang2018ACA}, distinguishes mild cognitive impairment (MCI, $n=38$) from NC ($n=37$). Third, the eyes open and eyes closed (EOEC) dataset~\cite{zhou2020toolbox}, collected from 48 college students (22 females, aged 19–31), is used for brain-state classification. 

All experiments were run on an NVIDIA A100 40G GPU using PyTorch3. The Adam optimizer started with a learning rate of \(1 \times 10^{-5}\), decaying by 0.5 every 50 epochs, with a weight decay of 0.03. The backbone GIN model uses three GNN layers, each with a two-layer MLP (hidden dimensions: [128, 256, 512]), followed by batch normalization and ReLU activation. The Brain3DCNN includes E2E3D layers (32 to 64 channels) and spatial-channel reduction modules. The model was trained for 100 epochs with a batch size of 32 and a dropout rate of 0.5. We set the matrix-based R\'enyi's $\alpha$-order entropy hyperparameters to $\sigma=5$ and $\alpha=1.01$~\cite{yu2021measuring,8787866}. Regularization coefficients $\beta_1$ and $\beta_2$ were tuned over $\{0, 0.0001, 0.001, 0.01, 0.1\}$ using tenfold cross-validation, with MvHo-IB achieving the best accuracy at $\beta_1=0.01$ and $\beta_2=0.1$. The fusion module combined both views via a 3-layer MLP with ReLU and dropout ($p=0.5$). All competing models were trained with their recommended hyperparameters.

\subsection{Experimental Results and Ablation Study}
We compare our proposed model with eight different methods, including three representative GNN models:
GCN~\cite{kipf2017semisupervised}, GAT~\cite{gat} and GIN~\cite{gin}, 
three state-of-the-art approaches based on information-theoretic principles:
SIB~\cite{9537601}, DIR-GNN~\cite{wu2022discoveringinvariantrationalesgraph} and BrainIB~\cite{BrainIB}, 
and two hypergraph-based approaches:
HYBRID~\cite{qiu2024learninghighorderrelationshipsbrain} and HMNet~\cite{li2025mhnetmultiviewhighordernetwork}, in three datasets in Table~\ref{tab:table1}. 
As can be seen, our approach consistently outperforms others by a significant margin, particularly on the UCLA and ADNI. BrainIB and hypergraph-based methods follow in ranking, leveraging information compression and HOIs, respectively, to enhance generalization. By naturally integrating both merits, our approach achieves the best performance.

\begin{table}[htbp]
\caption{Tenfold cross-validation performances of different models. The best performance is in bold, and the second-best is underlined.}
\label{tab:table1}
\renewcommand{\arraystretch}{1.2} 
\centering
\begin{tabular}{lccc}
\hline
\textbf{Method} & \textbf{UCLA} & \textbf{ADNI} & \textbf{EOEC} \\ \hline
GCN~\cite{kipf2017semisupervised}     & $62.27 \pm 6.21$ & $66.13 \pm 4.62$ & $70.92 \pm 8.56$ \\
GAT~\cite{gat}                        & $67.73 \pm 7.61$ & $66.28 \pm 8.69$ & $72.73 \pm 8.64$ \\
GIN~\cite{gin}                        & $65.91 \pm 8.21$ & $68.33 \pm 6.47$ & $75.41 \pm 9.65$ \\
DIR-GNN~\cite{wu2022discoveringinvariantrationalesgraph} & $75.72 \pm 8.37$ & $70.63 \pm 6.96$ & $80.12 \pm 6.21$ \\
SIB~\cite{9537601}                    & $72.76 \pm 8.13$ & $70.12 \pm 7.43$ & $80.42 \pm 7.97$ \\
BrainIB~\cite{BrainIB}                & $79.14 \pm 4.17$ & $\underline{72.47 \pm 5.32}$ & $82.06 \pm 5.43$ \\
HYBRID~\cite{qiu2024learninghighorderrelationshipsbrain} & $\underline{79.38 \pm 8.34}$ & $71.34 \pm 7.43$ & $81.97 \pm 7.43$ \\
MHNet~\cite{li2025mhnetmultiviewhighordernetwork}        & $79.22 \pm 6.72$ & $71.96 \pm 4.96$ & $\underline{82.87 \pm 5.43}$ \\
MvHo-IB                              & \textbf{$\boldsymbol{83.12 \pm 5.74}$} & \textbf{$\boldsymbol{73.23 \pm 4.37}$} & \textbf{$\boldsymbol{82.13 \pm 6.96}$} \\ \hline
\end{tabular}
\end{table}

To clarify the role of each module in our framework, we conduct an ablation study: Use only the GIN with pairwise FC measured with Pearson's correlation coefficient $\rho$. Combine GIN with nonlinear pairwise interaction measured with mutual information $I_\alpha$, while excluding the \(\mathcal{O}\)-information 3D tensor. Integrate the \(\mathcal{O}\)-information 3D tensor into the Brain3DCNN pathway. Enable both GIN and Brain3DCNN while incorporating the mutual information-based FC and the \(\mathcal{O}\)-information 3D tensor. Finally, include all components of MvHo-IB, together with the information bottleneck regularization, i.e., the last two terms in Eq.~\ref{eq5}.

\begin{table}[!ht]
\centering
\caption{Tenfold cross-validation results for the ablation study on three datasets. Bold indicates the best performance, while underlined denotes the second-best.
}
\label{tab:table2}
\renewcommand{\arraystretch}{1.2} 
\begin{tabular}{cccccc}
\hline
\textbf{Dataset} & $\rho$ & $I_\alpha$ &  $\mathcal{O}_\alpha$ & $I_\alpha + \mathcal{O}_\alpha$ (w/o IB)  & $I_\alpha +\mathcal{O}_\alpha$ (w. IB)  \\
\hline
UCLA & $65.91 \pm 8.21$ & $73.26 \pm 8.43$ & $74.29 \pm 5.43$ &  $\underline{82.51 \pm 5.96}$ & $\mathbf{83.12 \pm 5.74}$ \\
ADNI & $68.33 \pm 6.47$ & $69.23 \pm 7.29$ & $70.81 \pm 6.02$ & $\underline{72.21 \pm 6.10}$ & $\mathbf{73.23 \pm 4.37}$ \\
EOEC & $75.41 \pm 9.65$ & $76.52 \pm 6.61$ & $77.63 \pm 6.17$ & $\underline{81.34 \pm 6.92}$ & $\mathbf{82.13 \pm 6.96}$ \\
\hline
\end{tabular}
\end{table}

As shown in Table~\ref{tab:table2}, both the mutual information nonlinear pairwise interaction and the three–way $\mathcal{O}$–information HOI contain more discriminative information than simply using $\rho$. The multiview learning framework with IB regularization naturally fuses two kinds of complementary information, while effectively removing redundant information, thereby improving generalization.

\begin{figure}[htbp]
  \centering
  \begin{minipage}[b][6cm][c]{0.48\textwidth}
    \centering
    \captionof{table}{The top two interpretable pairwise and three-way interactions used by our model, identified with Grad-CAM. For EOEC, the top two pairwise groups are identical, only one group is presented.}
    \label{tab:table3}
    \vspace{1ex} 
    \resizebox{0.87\textwidth}{!}{%
\begin{tabular}{ccc}
\hline
\small \textbf{Dataset} & \small \textbf{pairwise} & \small \textbf{three-way HOI} \\ \hline
\small UCLA 
& \small \begin{tabular}[c]{@{}c@{}}HC-HC\\ HC-SC\end{tabular} 
& \small \begin{tabular}[c]{@{}c@{}}HC-HC-TP\\ HC-HC-SM\end{tabular} \\ 
\small ADNI 
& \small \begin{tabular}[c]{@{}c@{}}SMN-FPN\\ FPN-DMN\end{tabular} 
& \small \begin{tabular}[c]{@{}c@{}}SMN-FPN-FPN\\ SMN-FPN-CON\end{tabular} \\ 
\small EOEC 
& \small \begin{tabular}[c]{@{}c@{}}SMN-FPN\\ \end{tabular} 
& \small \begin{tabular}[c]{@{}c@{}}SMN-FPN-FPN\\ FPN-DMN-CON\end{tabular} \\ \hline
\end{tabular}
    }
  \end{minipage}\hfill
  \begin{minipage}[b][6cm][c]{0.48\textwidth}
    \centering
    \includegraphics[width=\textwidth,height=6cm,keepaspectratio]{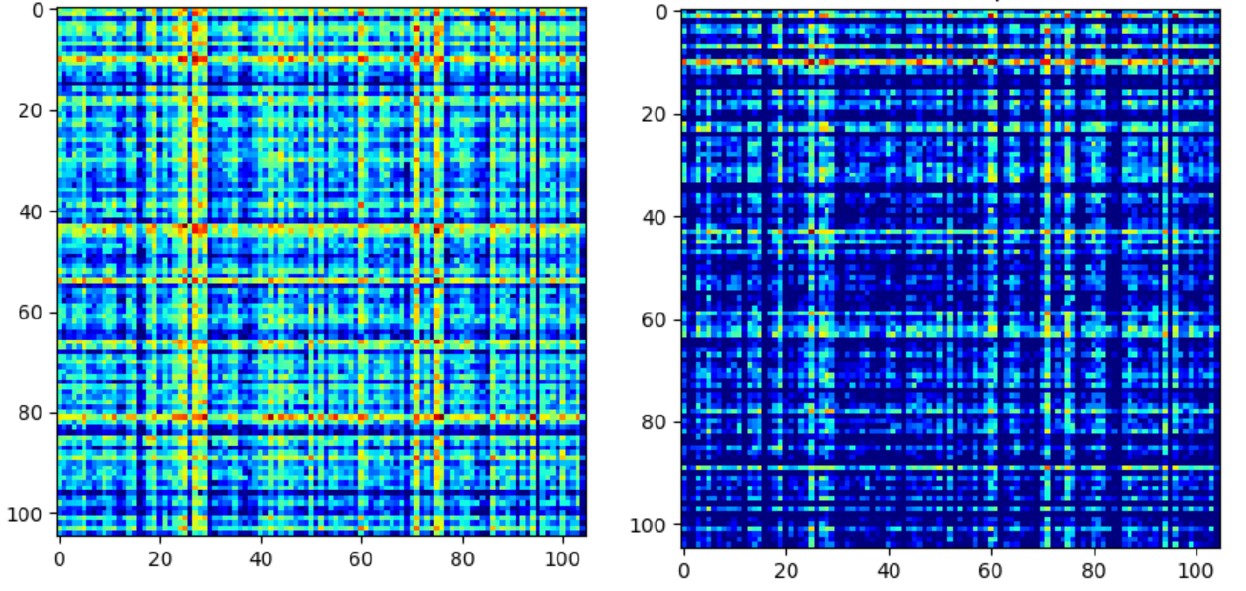}
    \captionof{figure}{Heatmap visualization of interpretable connectivities by Grad-CAM in UCLA dataset. Left: pairwise interactions; Right: one slice of 3D tensor interactions.}
    \label{fig:miccai3}
  \end{minipage}
\end{figure}

In order to validate our model's interpretability, we applied Grad-CAM~\cite{grad-cam} to three datasets as shown in Figure~\ref{fig:miccai3}. The abbreviations are: sensorimotor network (SMN), occipital network (ON), fronto-parietal network (FPN), default mode network (DMN), cingulo-opercular network (CON), cerebellum network (CN), Higher Cognition network (HC), Subcortical network (SC), Temporal network (TP), and Sensorimotor network (SM). We generated pairwise and three-way Grad-CAM heatmaps in two input views to identify the discriminative interactions distinguishing healthy individuals from those with mental illness. Table~\ref{tab:table3} shows informative pairwise interactions within HC and between HC and SC in the UCLA schizophrenia dataset, aligning with findings from previous studies~\cite{li2023aberranthighorderdependenciesschizophrenia}.
Moreover, our model revealed significant and novel triple-network interactions among HC, TP, and SM, beyond pairwise associations, underscoring the added value of triple-network analysis for future research and validation.

\section{Conclusions and Future Work}

In this work, we propose a novel and effective framework that leverages HOIs to enhance diagnostic accuracy in fMRI-based mental disorder classification. Our method serves as a principled alternative to hypergraphs. We validate our framework on three benchmark fMRI datasets and compare it against eight competitive methods, confirming its superior precision. Moreover, interpretability analyses verify the reliability of our approach by revealing neurobiologically plausible three–way HOI biomarkers that offer new and promising insights into the distributed neural mechanisms underlying psychiatric conditions. 

While this paper focuses on third-order $\mathcal{O}$-information, the formulation naturally extends to higher orders, yielding a $K$-way tensor~\cite{li2023aberranthighorderdependenciesschizophrenia}. However, computing higher-order $\mathcal{O}$-information is computationally demanding. Two promising directions for scalability are: (1) leveraging an analytical form under Gaussian assumptions~\cite{li2023aberranthighorderdependenciesschizophrenia}, and (2) using low-rank approximations of the matrix-based entropy functional~\cite{10077595}.

\begin{credits}
\subsubsection{\ackname} This study was funded in part by the Research Council of Norway (RCN) under grant 309439.

\subsubsection{\discintname}
No competing interests.

\end{credits}
%
%
%
%
\bibliographystyle{plain}
\bibliography{Paper-0646}

\end{document}